\documentclass[11pt]{article}

\PassOptionsToPackage{table,svgnames,dvipsnames}{xcolor}
\usepackage[preprint]{acl}

\usepackage{times}
\usepackage{latexsym}
\usepackage[T1]{fontenc}
\usepackage[utf8]{inputenc}

\usepackage{microtype}

\usepackage{inconsolata}

\usepackage{graphicx}

\usepackage{float}
\usepackage{booktabs}
\usepackage{svg}
\newcommand{\orcid}[1]{\href{https://orcid.org/#1}{\includesvg[width=10pt]{orcid}}}
\usepackage{mwe}
\usepackage{scrhack} % necessary for listings package
\usepackage{xcolor}
\usepackage{listings}
\usepackage{lstautogobble}
\usepackage{tikz}
\usepackage{tabularx}
\usepackage{array}
\usepackage{amsmath}
\usepackage[dvipsnames]{xcolor}
\usepackage{multirow}
\usepackage{siunitx}
\definecolor{codegreen}{rgb}{0,0.6,0}
\definecolor{codegray}{rgb}{0.5,0.5,0.5}
\definecolor{codepurple}{rgb}{0.58,0,0.82}
\definecolor{backcolour}{rgb}{0.95,0.95,0.92}

\lstdefinestyle{mystyle}{
    backgroundcolor=\color{backcolour},   
    commentstyle=\color{codegreen},
    keywordstyle=\color{magenta},
    numberstyle=\tiny\color{codegray},
    stringstyle=\color{codepurple},
    basicstyle=\ttfamily\footnotesize,
    breakatwhitespace=false,         
    breaklines=true,                 
    captionpos=b,                    
    keepspaces=true,                 
    numbers=left,                    
    numbersep=5pt,                  
    showspaces=false,                
    showstringspaces=false,
    showtabs=false,                  
    tabsize=2
}

\title{AutoJourn: Multi-Perspective Summarisation, Bias Detection and Bias Neutralisation for LLM-Generated News in Automated Journalism}

\author{
  \textbf{Himel Ghosh\textsuperscript{1,2}},
  \textbf{Ahmed Mosharafa\textsuperscript{1}},
  \textbf{Georg Groh\textsuperscript{1}}
  \\
  \textsuperscript{1}Technical University of Munich, Germany,
  \textsuperscript{2}Sapienza University of Rome, Italy
  \\
  \small{
    \textbf{Correspondence:} \href{mailto:himel.ghosh@tum.de}{himel.ghosh@tum.de}
  }
}

\begin{document}
\maketitle
\begin{abstract}
We present \textsc{AutoJourn}, a demonstration system for multi-perspective news generation and bias-aware evaluation using large language models (LLMs). The system tackles three core challenges in responsible automated journalism: extracting diverse perspectives from unstructured social media discussions, generating summaries that preserve viewpoint diversity, and detecting or mitigating bias in AI-generated news. The pipeline integrates advanced prompt engineering with optional retrieval augmentation to produce semantically diverse perspective sets, a multi-perspective summarisation module that merges conflicting viewpoints into balanced summaries, and a bias analysis suite supporting sentence-level bias detection and type classification in the generated news article, and automatic neutralisation. Users can inspect perspective clusters, compare stance-specific summaries, generate news articles, and apply bias-aware rewrites directly in the interface. We evaluate each component with intrinsic metrics—semantic diversity, summary quality, and bias reduction and show improvements over strong baselines while maintaining content fidelity. A live, publicly accessible demo accompanies the paper to facilitate reproducibility and further research on socially responsible automated journalism.
\end{abstract}

\section{Introduction}

Since the introduction of Transformer architectures~\cite{Vaswani2017}, large language models (LLMs)~\cite{app15116110,brown2020language,bommasani2021foundation} have enabled fluent, domain-general text generation and are now widely explored in automated journalism~\cite{graefe2016guide,diakopoulos2019automating}. While such systems offer scalability, they also risk flattening diverse viewpoints into a single narrative and reproducing subtle biases present in source data or model training signals. This is particularly problematic when public discourse is increasingly shaped by social media discussions~\cite{kwak2010twitter,zubiaga2018discourse}, where opinions are fragmented, polarized, and influenced by algorithmic filtering and selective exposure~\cite{pariser2011filter,gentzkow2006media,tversky1974judgment,cinelli2021echo,del2016echo,sunstein2017republic}.

\begin{figure}[H]
    \centering
    \includegraphics[width=\linewidth]{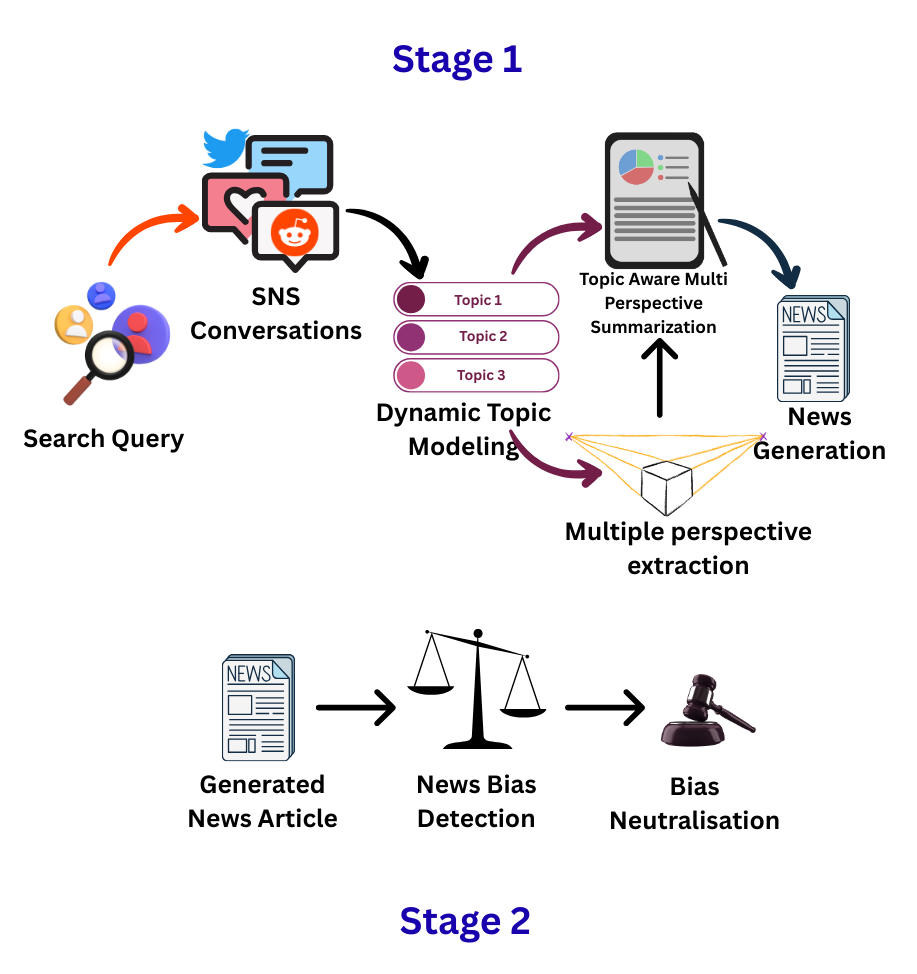}
    \caption{System pipeline: multi-perspective extraction, stance-specific and merged summarisation, perspective-conditioned news generation, and bias-aware evaluation and neutralisation.}
    \label{fig:pipeline}
\end{figure}

Existing NLP tools often target isolated components of this challenge. Summarisation systems, even controllable ones~\cite{fan2018controllable,goyal2021evaluating}, typically merge competing viewpoints into one “neutral’’ account. Work on bias detection and mitigation~\cite{sheng2019woman,bender2021dangers,spinde2021babe,pryzant2020automatically} identifies problematic framing but is rarely integrated into end-to-end news generation. As a result, current automated journalism pipelines lack support for (i) surfacing multiple perspectives, (ii) generating viewpoint-aware articles, and (iii) evaluating the bias of the resulting text.

In this demonstration paper, we present an integrated system that addresses these gaps by combining \textbf{multi-perspective extraction and summarisation} with \textbf{bias evaluation} for automated news writing. Given raw input text (e.g., a social media conversation), the system:  
(i) extracts topics and associated stances using LLM-based perspective identification;  
(ii) produces perspective-specific and merged multi-perspective summaries;  
(iii) generates full news articles conditioned on a chosen topic and corresponding multi-perspective summary; and  
(iv) evaluates the generated article using a neural bias detector and bias-type classifier, and produces a bias-neutralised version.

% This system demo focuses on the two core stages: perspective-aware generation and bias-sensitive evaluation. The resulting prototype demonstrates how automated journalism systems can be made more transparent, diverse in viewpoint representation, and socially responsible.
A key novelty of this work lies in offering a unified, end-to-end framework that connects perspective extraction, multi-perspective summarisation, controlled news generation, and bias evaluation within a single demonstrable system. While prior research has examined these components in isolation, our prototype is, to our knowledge, the first to operationalize them jointly for automated journalism. This integration enables users to trace how viewpoints propagate from source discussions into generated narratives and to assess the bias of those narratives through explicit, model-driven analysis. By combining interpretability, viewpoint diversity, and bias mitigation, the system illustrates a practical path toward more transparent and socially accountable applications of LLMs in news production.

\section{Related Work}

Automated journalism has been explored both commercially and academically, with systems such as Reuters’ Lynx Insight and The Washington Post’s Heliograf demonstrating how AI can support story drafting and data-driven reporting~\cite{reuters_lynx_insight_2018,lynx_insight_platform,wapo_heliograf_2016,heliograf_case_study_2025}. Research prototypes extend these ideas by generating summaries of political debates or producing “balanced’’ narratives, though these systems often rely on single-perspective summaries that overlook viewpoint diversity.

Topic modeling plays an important role in organising large conversational inputs. Recently, prompt-based topic modeling~\cite{wang2023promptinglargelanguagemodels,mu2024largelanguagemodelsoffer,pham2024topicgptpromptbasedtopicmodeling} has emerged as a flexible alternative, leveraging LLMs to produce coherent, interpretable topics without corpus-level training. These methods are particularly suited to dynamic social-media content and align with the prompt-driven topic extraction used in our system.

The rise of LLMs has intensified concerns about bias, fairness, and representational imbalance in AI-generated news. Recent studies show that LLM-generated articles exhibit systematic ideological and topical biases~\cite{llm_bias_nature_2024}, reflecting both training data disparities and model reasoning patterns. Surveys on fairness in LLM-generated content highlight three core dimensions: representational, allocational, and procedural fairness, which are critical for evaluating automated news systems~\cite{bias_fairness_survey_2024}. Complementary efforts in bias detection and mitigation, including neural media-bias classifiers such as BABE~\cite{spinde2021babe} and neutrality style transfer approaches using the Wiki Neutrality Corpus~\cite{pryzant2020automatically}, provide tools for identifying or rewriting biased passages, but they are typically applied in isolation rather than integrated into generation workflows.

Our work also draws on advances in stance detection and multi-perspective summarisation (MPS) which aims to represent diverse or conflicting viewpoints rather than collapse them into a single narrative~\cite{fabbri2021,yadav2022}. Frameworks such as PerSphere~\cite{luo2024} retrieve opposing documents and generate perspective-specific summaries, while LLM-driven methods leverage prompt engineering, fine-tuning, and knowledge distillation to produce viewpoint-aware summaries~\cite{hayati2024,zhang2025,aly2025}. Multi-agent debate systems further improve diversity and accuracy by combining complementary model outputs~\cite{ki2025}. These approaches show strong promise for capturing distinct stances, but they seldom connect perspective extraction to downstream news writing or subsequent bias evaluation.

Finally, work on bias mitigation in generated text investigates how to rewrite content into more neutral phrasing while preserving factual content~\cite{pryzant2019}. Despite progress, challenges remain due to the subjective and context-dependent nature of bias~\cite{ghosh2025} and the difficulty of balancing neutrality with informativeness.

Our system bridges several of these lines of research by jointly integrating (i) multi-perspective extraction, (ii) prompt-based topic induction, (iii) perspective-based summarisation, (iv) controlled news generation, and (v) sentence-level bias analysis and neutralisation into one end-to-end demonstrable pipeline for automated journalism.

\section{System Description}
\label{sec:system-description}

Given a social-media discussion, our system identifies diverse viewpoints, produces stance-specific and merged summaries, generates a full news article conditioned on these perspectives, and finally evaluates and neutralises bias in the generated text. Figure~\ref{fig:pipeline} illustrates the overall workflow.

\begin{figure}[H]
    \centering
    \includegraphics[width=1\linewidth]{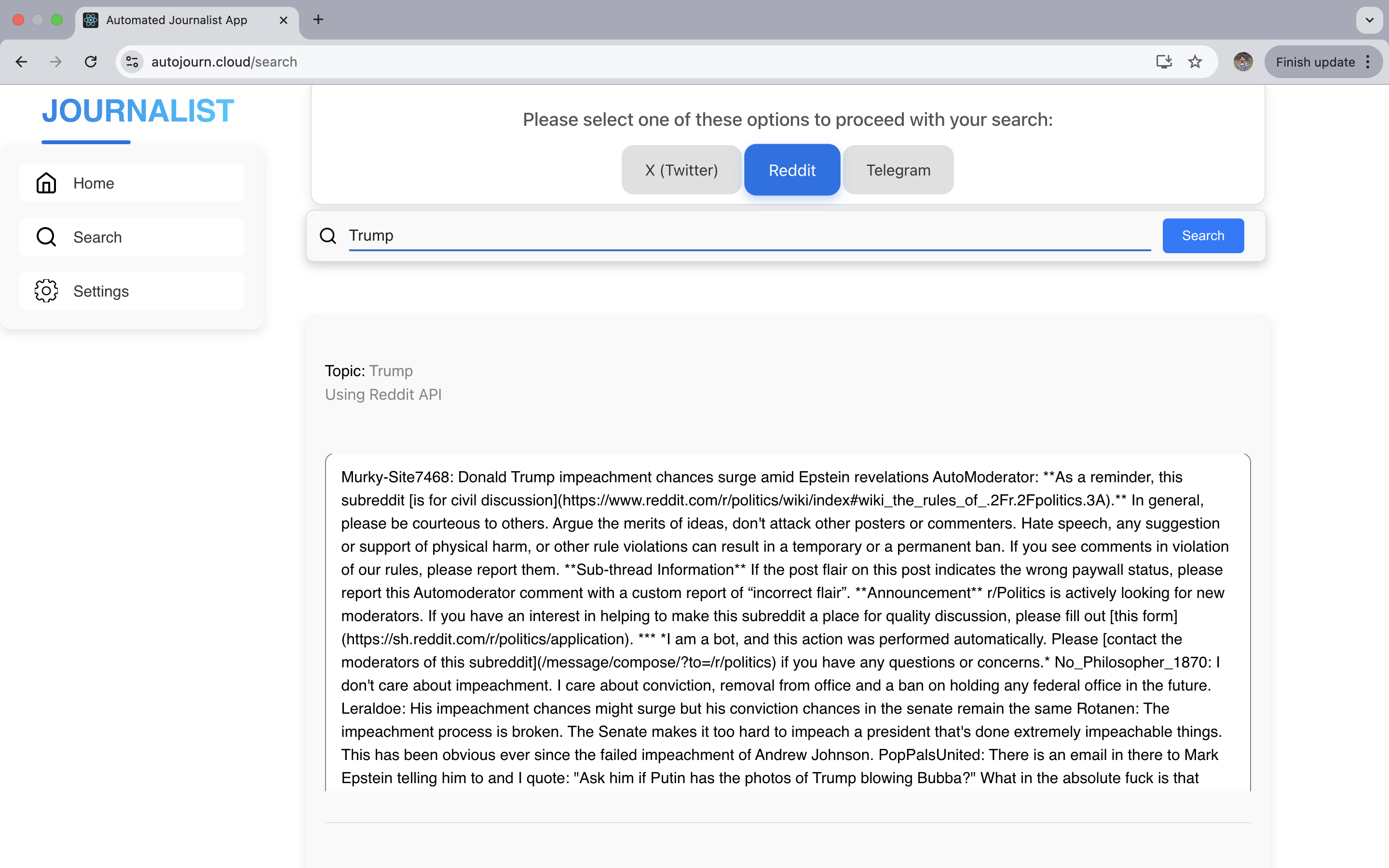}
    \caption{Screenshot of the AutoJourn App from the website. It shows the search results after a query search.}
    \label{fig:autojournss}
\end{figure}

\subsection{Application Demonstration}

Users can access the system through a desktop browser at \url{https://autojourn.cloud/}. The homepage allows users to submit a query, which by default retrieves relevant Reddit discussions. After selecting a thread, the user proceeds to the topic-modelling interface by clicking \textit{Get Topics \& Perspectives}. The system then displays the generated topics alongside their associated multiple perspectives.
Selecting a topic reveals the extracted \textit{Agree} and \textit{Disagree} perspectives. By clicking \textit{Summarize}, the system produces stance-specific summaries and a merged, multi-perspective summary. Choosing \textit{Generate News} creates a full news article conditioned on the selected topic and preferred writing style.
On the news page, the \textit{Detect Bias} function highlights biased sentences, with hover tooltips showing predicted bias type and confidence. The \textit{Neutralise Bias} option provides rewritten neutral alternatives, which can be inserted directly into the article via the \textit{Replace Bias} button.
Users may also export a comprehensive PDF report via \textit{Full Report}, containing the source conversation, extracted perspectives, summaries, generated article, bias annotations, and neutralised versions. Figure~\ref{fig:autojournss} provides an overview, with additional screenshots in Appendix~\ref{screens}. A full demonstration video is available at: \url{https://youtu.be/XZWGlWqhJHQ}.

\subsection{Topic Modeling}
For topic extraction, AutoJourn implements a prompt-based topic modeling pipeline inspired by recent LLM-driven approaches~\cite{wang2023promptinglargelanguagemodels,mu2024largelanguagemodelsoffer,pham2024topicgptpromptbasedtopicmodeling}.
Given a conversation thread or document, a structured zero-shot prompt instructs the LLM to identify the top five semantically coherent topics, estimate their relative proportions, and list 5 representative keywords per topic. The Appendix~\ref{dtmprompt} provides the exact template used.

\subsection{Multi-Perspective Extraction}

This stage identifies distinct viewpoints present in the source text. Following prior work on perspective elicitation~\cite{hayati2024}, we use a GPT-4o model prompted to generate multiple \emph{Agree} and \emph{Disagree} stances with concise, criteria-based justifications. This enables the model to produce opinions that differ not only in polarity but also in underlying value dimensions.

To improve diversity and contextual grounding, the system optionally augments the input using a lightweight retrieval-augmented generation (RAG) mechanism~\cite{lima2025,logé2025}. Given a topic or a statement extracted from the text, we retrieve semantically related documents using TF--IDF similarity and sentence-embedding reranking (e.g., e5-large-v2). Retrieved snippets provide additional argumentative context, supporting richer and more specific perspective generation, in line with evidence-based prompting frameworks used in multi-perspective summarisation~\cite{luo2024}.

The final output of this stage is a structured dictionary of perspectives for each topic:
\[
\begin{array}{c}
   \mathcal{P}[t] = \{\texttt{Agree}: p_1^A,\dots,p_k^A;\;\\
\texttt{Disagree}: p_1^D,\dots,p_k^D\}.
\end{array}
\]

See Appendix-\ref{mpePrompt}, \ref{RAGPrompt} for the prompt designs.

\subsection{Multi-Perspective Summarisation}

Building on the PerSphere framework~\cite{luo2024}, the next stage converts extracted perspectives into stance-specific summaries. For each topic, the system generates an \textit{Agree} summary and a \textit{Disagree} summary using prompt-guided abstractive summarisation. 
%Summaries are required to (i) remain grounded in the original perspectives, (ii) avoid overlap between stance-specific content, and (iii) preserve argumentative structure, echoing principles from argument mining~\cite{khatib2021}.

After producing the two stance-specific summaries, the system generates a \textit{Merged} summary that synthesises both viewpoints using neutral, balanced language. Inspired by SubSumm~\cite{jiang2023} and multi-document summarisation research~\cite{kouris2021}, the merging prompt emphasises equal representation, removal of emotionally charged phrases, and inclusion of evidence from both sides. See Appendix-\ref{mpsPrompt} for the prompts.

The result is a triplet \((A, D, M)\): an agree-focused summary, a disagree-focused summary, and a merged multi-perspective summary.

\subsection{News Article Generation}

The topic and merged summary then serve as conditioning signals for generating full news articles. Using GPT-4o, the system expands each summary into a coherent news article while preserving the intended perspective framing. Check Appendix-\ref{NGPrompt} for the prompt design. Users can generate a \emph{Balanced/Neutral} article based on the merged summary.

This mirrors controllable generation approaches used in political and opinion-aware summarisation~\cite{fan2018controllable,goyal2021evaluating}.

% \begin{figure}[H]
%     \centering
%     \includegraphics[width=1\linewidth]{Figs/Image19.png}
%     \caption{Bias Detector Architecture: Roberta Base Finetuned on BABE Dataset}
%     \label{fig:biasdetector}
% \end{figure}

\subsection{Bias Detection}

To evaluate the generated article, we incorporate a sentence-level bias analysis module \cite{ghosh2025} grounded in prior work on media bias~\cite{spinde2021babe,pryzant2020automatically}. The system uses a fine-tuned RoBERTa classifier trained on the BABE dataset~\cite{spinde2021babe} to detect biased sentences. Each sentence is assigned (i) a bias probability and (ii) a predicted bias type using the GusNet classifier~\cite{powers2025}, covering political, demographic, or socioeconomic categories. \citet{ghosh-werner-2026-llm} uses this pipeline for real-time comparison of biases in LLM outputs.

This produces a transparent, interpretable bias map over the generated article.

\subsection{Bias Neutralisation}

When biased sentences are identified, the system applies a two-stage mitigation pipeline inspired by the WNC neutrality work~\cite{pryzant2019}. First, a fine-tuned encoder--decoder model (BART or T5) rewrites biased sentences into neutral formulations while preserving meaning. Second, if the model fails to sufficiently neutralise a sentence, a fallback GPT-based rewrite is triggered using prompt shown in Appendix-\ref{BNPrompt}, following recent LLM-based style-transfer work~\cite{reif2022,furniturewala2024}. This ensures that even difficult cases receive a neutralised variant.

\begin{figure}[H]
    \centering
    \includegraphics[width=1\linewidth]{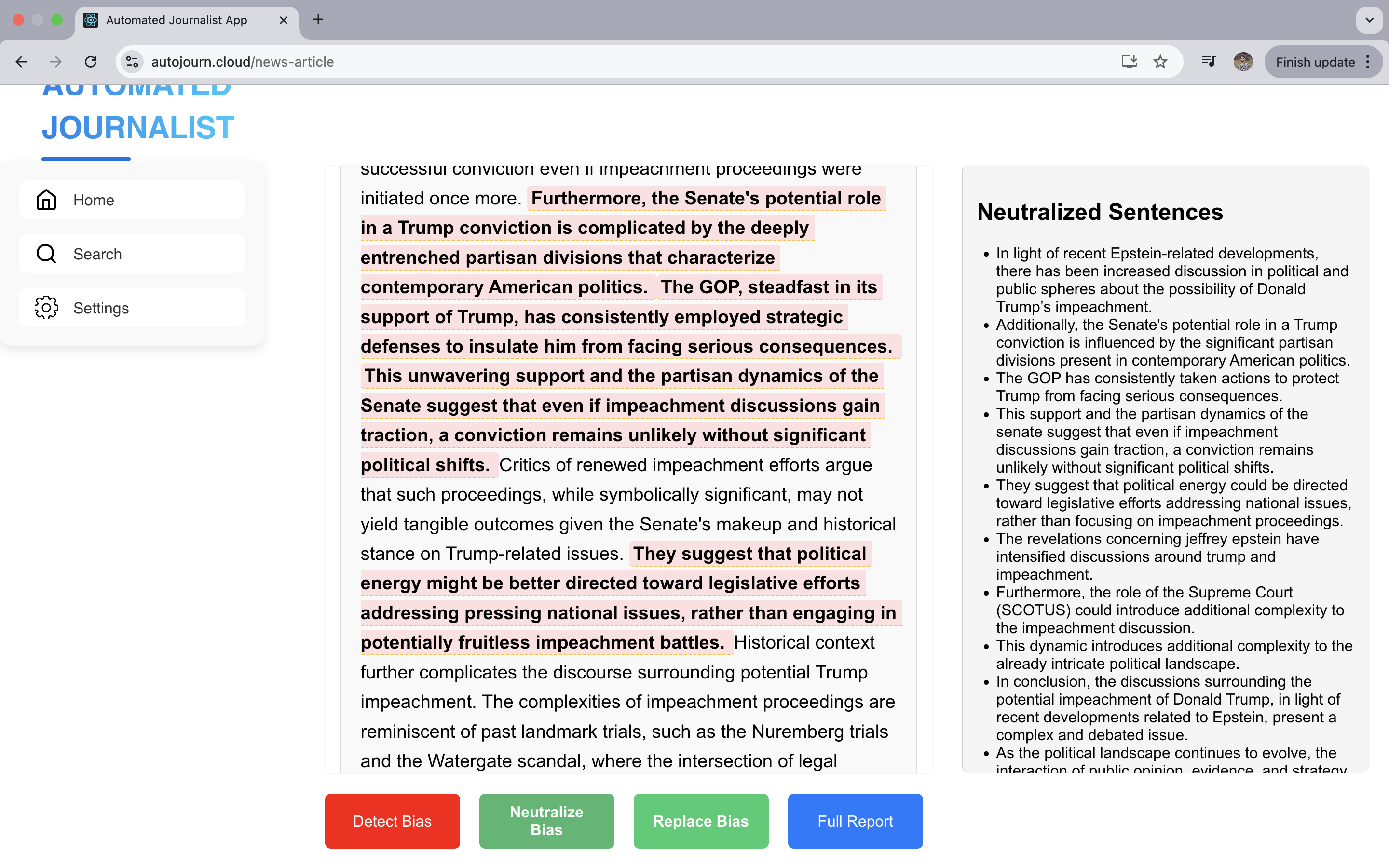}
    \caption{Application Screenshot: Demonstration of the highlighted biased sentences in the generated news and the neutralised alternatives.}
    \label{fig:autojournss2}
\end{figure}

The result is a fully neutralised version of the generated news article, enabling direct comparison between the original output and a bias-mitigated alternative. See Fig. \ref{fig:autojournss2} for reference.

While the individual components draw on existing techniques, the contribution of AutoJourn lies in their unified orchestration into a transparent, end-to-end workflow for multi-perspective news generation. Previous systems typically address only one stage (e.g., summarization or bias detection), whereas AutoJourn provides an integrated pipeline with actionable user-facing controls and bias-aware rewriting.

\subsection{Intended Audience}

\textsc{AutoJourn} is intended for journalists, NLP researchers, and practitioners in computational social science. For researchers, it offers a modular testbed for studying multi-perspective generation, bias detection, and controllable news rewriting. For journalists and newsroom technologists, the system provides practical tools to inspect perspective diversity, compare stance-specific summaries, and highlight biased sentences in AI-generated drafts, supporting transparent and efficient editorial workflows. Overall, \textsc{AutoJourn} serves as a unified platform for exploring responsible automated journalism.

\section{Evaluation}
\label{sec:evaluation}

We evaluate the system along four axes that mirror its main components:
(i) multi-perspective extraction, (ii) multi-perspective summarisation,
(iii) sentence-level bias detection, and (iv) bias neutralisation.
Our goal is not to exhaustively benchmark each sub-module, but to show
that the integrated system produces diverse perspectives, balanced summaries,
and effectively detects and reduces bias in generated news. Detailed
metrics, plots, and screenshots are provided in the appendices.

\subsection{Multi-Perspective Extraction}

Following prior work on diversity extraction from LLMs~\cite{hayati2024},
we evaluate our perspective extraction module on two subsets:
(i) \textbf{SocialChem-101} ~\cite{socialchem}), and
(ii) \textbf{CMV} \cite{cmv}.
For each input, GPT-4o generates 6 opinions (Agree/Disagree) using
our final criteria-based prompt, optionally with RAG context.

We adopt the semantic diversity metric of Hayati et al.~\cite{hayati2024},
computing pairwise cosine distances between sentence embeddings of
generated reasons. Across both datasets, our GPT-4o variants achieve
the highest semantic diversity scores among all models reported in
their study. See Fig \ref{fig:semdiv-combined} for the results. Further evaluation details are in Appendix~\ref{evalres} Table \ref{tab:semdiv_criteria_combined}.

\begin{figure}[H]
    \centering
    \caption{Semantic diversity (criteria-based prompting). 
    Top: radar plots comparing all models for Social-Chem-101 (left) and CMV (right). 
    Bottom: line plot (left) and grouped bar chart (right) provide complementary views. 
    Our variants are GPT-4o with and without RAG.}
    \includegraphics[width=.48\linewidth]{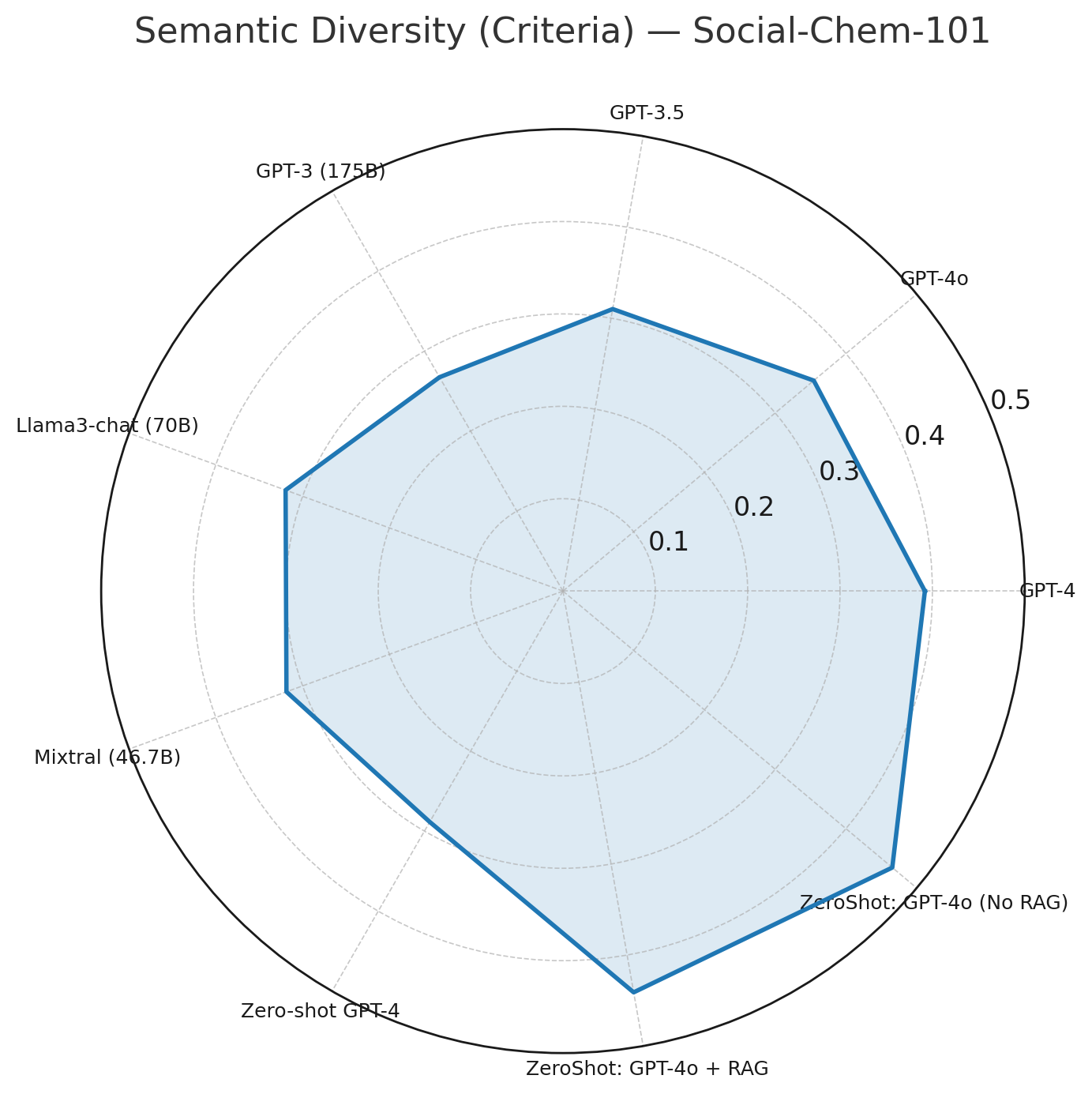}\hfill
    \includegraphics[width=.48\linewidth]{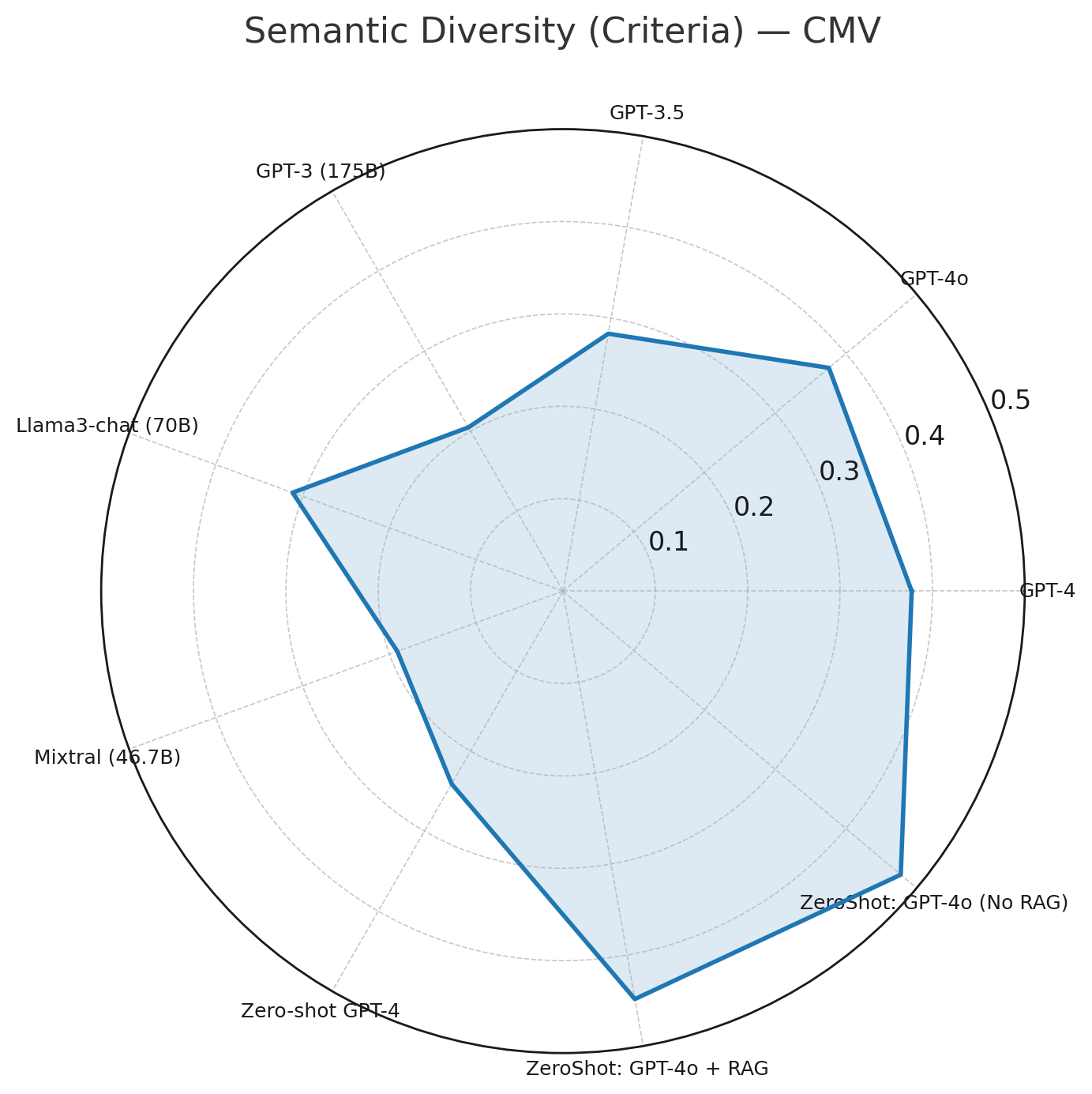}\\[3pt]
    \includegraphics[width=.48\linewidth]{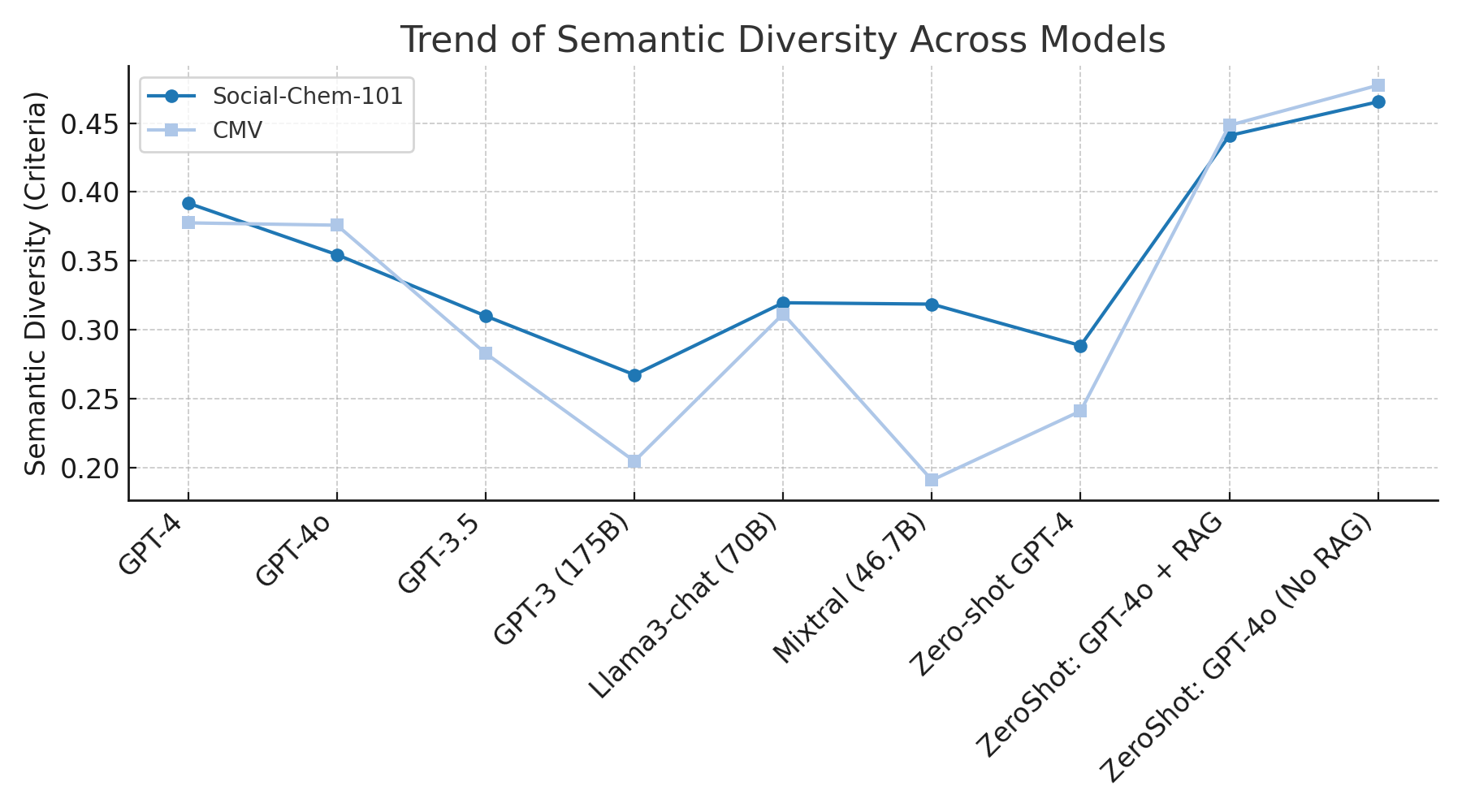}\hfill
    \includegraphics[width=.48\linewidth]{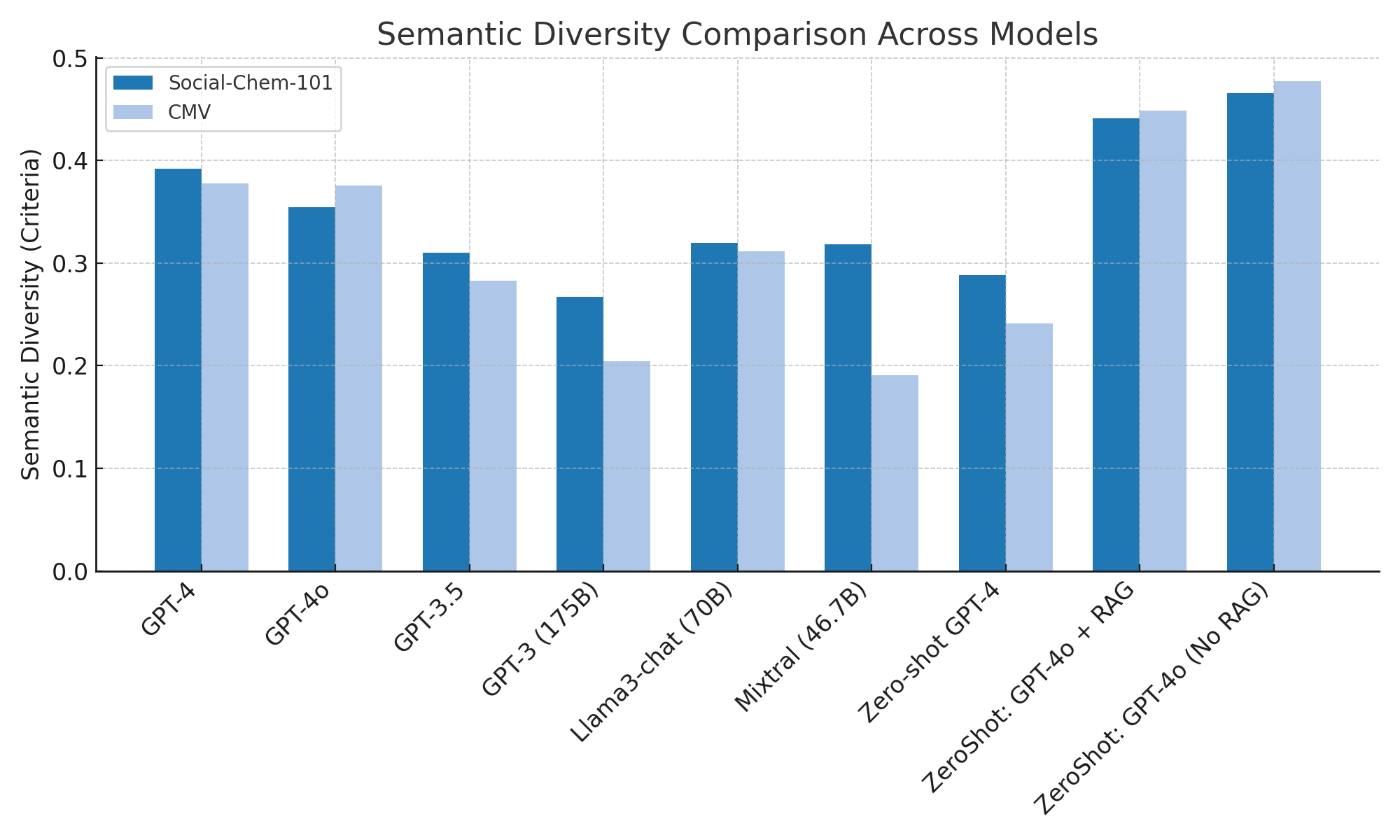}
    \label{fig:semdiv-combined}
\end{figure}

Interestingly, the RAG-augmented variant shows slightly lower raw
diversity than our designed prompting. Hence, in practice,
we use the GPT-4o prompt in our demo system.

\subsection{Multi-Perspective Summarisation}

For each topic, the system generates (i) an \textit{Agree} summary, (ii) a \textit{Disagree} summary, and (iii) a \textit{Merged} summary intended to concisely represent both sides. Since no established benchmark exists for multi-perspective summarisation with stance preservation, our evaluation focuses on demonstrating how summaries can be inspected within the system, rather than establishing new state-of-the-art results.

To obtain a structured, model-based assessment of summary quality, we employ an LLM-as-judge protocol inspired by Luo et al.~\cite{luo2024}. The prompt (Appendix~\ref{GPTEvalPrompt}) guides the evaluator to examine faithfulness to the input, coverage of perspectives, and distinctiveness between stances. This provides a qualitative check on whether the summaries adhere to the intended task definition.

We additionally report surface-level fluency scores using Grammarly—not as a proxy for content accuracy or perspective preservation, but as an auxiliary readability indicator for users of the demo system. These fluency scores should not be interpreted as a comprehensive evaluation metric for summarisation quality. Table~\ref{tab:main-summary-results2} reports these values for completeness.

\begin{table}[H]
\centering
\small
\caption{Grammarly evaluation results for summaries generated by the system (mean $\pm$ 95\% CI). Concat is a naive concatenation of the agree and disagree summary compared with the merged summary}
\label{tab:main-summary-results2}
\begin{tabular}{lc}
\toprule
System & Grammarly $\uparrow$ \\
\midrule
Agree (stance)& 90.2 $\pm$ 4.2 \\
Disagree (stance)   & 93.5 $\pm$ 3.0 \\
Concat (Agree+Dis.) & 88.1 $\pm$ 5.0 \\
\textbf{Merged} & \textbf{85.7 $\pm$ 3.4} \\
\bottomrule
\end{tabular}
\end{table}

As expected, merging opposing viewpoints into a single compact paragraph slightly reduces surface fluency relative to stance-specific summaries. However, this trade-off is acceptable in our demonstration context, where the goal is to provide users with accessible mechanisms for inspecting how multiple perspectives are distilled into a single narrative.

\subsection{Bias Detection}

Our system requires a reliable sentence-level bias detection to flag potentially biased spans in generated news articles. We integrate an existing RoBERTa-based bias detector introduced in prior work~\cite{ghosh2025}, publicly released on Hugging Face.\footnote{\url{https://huggingface.co/himel7/bias-detector}}

This detector was originally trained and evaluated on the BABE dataset~\cite{spinde2021babe} under the standard protocol of Krieger et al.~\cite{Krieger2022}, and has been evaluated against the DA-Roberta Baseline of Krieger et al. \cite{Krieger2022}. Our contribution  to demonstrate how such a detector can be embedded into an interactive multi-perspective journalism pipeline.

% \begin{figure}[H]
% \centering
% \includegraphics[width=0.75\linewidth]{Figs/kfoldval.png}
% \caption{Macro F1 scores reported in prior work for the bias detector used in our system. We do not reproduce or extend these results; the detector is used as an off-the-shelf component.}
% \label{fig:fold-performance}
% \end{figure}

In our system, the detector provides sentence-level bias probabilities that drive several user-facing features: highlighted biased spans in generated articles, click-to-inspect bias types and confidence. For bias-type explanations, we integrate the complementary classifier of~\cite{powers2025}, allowing users to view the predicted category (e.g., political, racial, socioeconomic).

We are demonstrating how such a detector can be integrated into a multi-perspective news generation pipeline and how users can interact with the resulting bias signals during the authoring and editing process.

\subsection{Bias Neutralisation}

To support users in editing biased or subjective passages in generated news, our system incorporates existing bias-neutralisation models trained on the WNC corpus~\cite{pryzant2019}. We use the neutralisers (BART \footnote{\url{https://huggingface.co/himel7/bias-neutralizer-bart}} and T5-small \footnote{\url{https://huggingface.co/himel7/bias-neutralizer-t5s}}) and a
pipeline that combines BART with an LLM fallback neutraliser, as
described in \S\ref{sec:system-description}. The neutralisation of the biased sentences is triggered when the user clicks on the "Neutralise Bias" button. Evaluation is performed
on a held-out WNC split plus a small set of biased sentences from
BABE and our generated news. We measure (i) content preservation via
BLEU/ROUGE, and (ii) \emph{bias reduction} as the relative decrease
in bias probability from the detector before and after rewriting.

\begin{table}[H]
\centering
\begin{tabular}{lccc}
\hline
\textbf{Model} & \textbf{Bias Reduction (\%)} \\
\hline
BART Neutraliser & 84.7 \\
T5-Small Neutraliser & 82.1 \\
BART + LLM & \textbf{91.5} \\
\hline
\end{tabular}
\caption{Bias neutralization performance on combined WNC and test set.}
\label{tab:neutralization-results2}
\end{table}
Both encoder–decoder models achieve high similarity with neutral
references (ROUGE-L $\approx$ 0.96) while substantially decreasing
bias probabilities (82–85\% relative reduction). The BART + LLM
fallback pipeline yields the strongest bias reduction (about 92\%)
with only a minor drop in lexical similarity, confirming that the
fallback is beneficial for difficult cases where the fine-tuned model
fails to sufficiently edit the sentence. See Table ~\ref{tab:neutralization-results2} for results and Appendix \ref{evalres} Table \ref{tab:neutralization-results} for more detailed results.
In the demo, users can toggle between the original and neutralised article, visually inspect
highlighted changes, and export a bias report combining detection,
bias types~\cite{powers2025}, and neutralised text.

% Finally, we note a limitation: bias-neutralization evaluation partially relies on this same detector, introducing some metric circularity. To mitigate this, we complement detector-based metrics with human judgments and example-based analysis. Overall, the strong empirical performance and improved contextual sensitivity justify using our detector as the primary bias signal in the AutoJourn pipeline.

These results show that our system (i) extracts
diverse, grounded perspectives, (ii) produces stance-aware summaries, and (iii) reliably
detects and reduces bias in generated news articles.

\begin{figure}[H]
    \centering
    \includegraphics[width=1\linewidth]{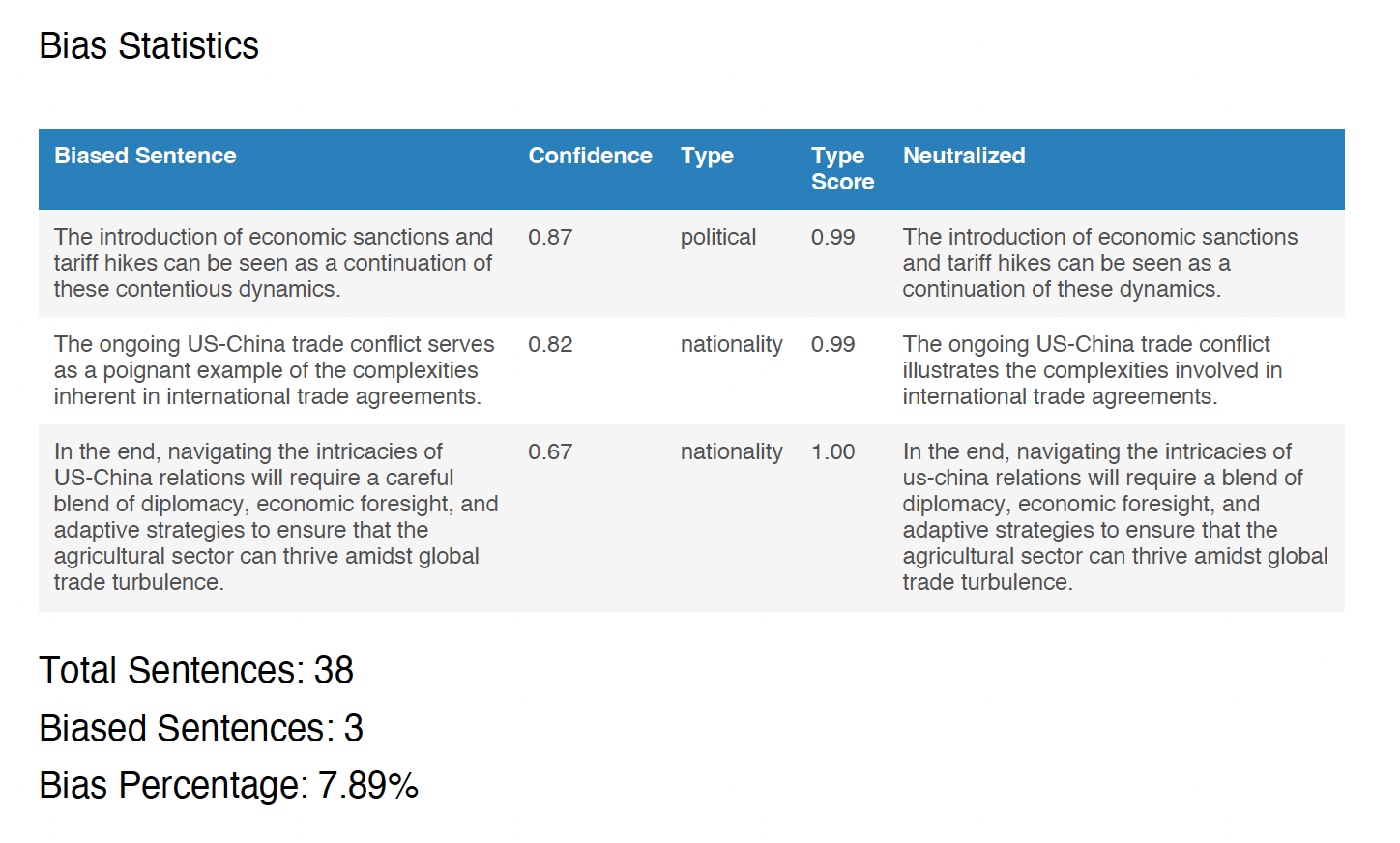}
    \caption{Screenshot of the Final Report downloaded from the application. It demonstrates the Bias Statistics present in a generated news article of this application}
    \label{fig:report}
\end{figure}

As a final step in the bias analysis workflow, users may download a complete Bias Report that compiles all bias-related outputs into a single, audit-ready document. The report aggregates: the original extracted conversation, detected biased sentences with confidence scores, predicted bias types and classifier probabilities, neutralised rewrites for each biased sentence, bias distribution visualisations (pie charts and statistics), and the fully neutralised version of the generated article. A screenshot of the report is given in Fig. \ref{fig:report}.

\textbf{Availability \& Licensing.}
The AutoJourn system is available as an online demonstration at \url{https://autojourn.cloud} for research and evaluation purposes. 
The underlying source code is not publicly released and can be distributed under a research-only usage agreement.


\bibliography{custom}

@article{app15116110,
AUTHOR = {Zhang, Jinyi and Iglesias, Carlos Á.},
TITLE = {Special Issue on Recent Applications of Machine Learning in Natural Language Processing (NLP)},
JOURNAL = {Applied Sciences},
VOLUME = {15},
YEAR = {2025},
NUMBER = {11},
ARTICLE-NUMBER = {6110},
URL = {https://www.mdpi.com/2076-3417/15/11/6110},
ISSN = {2076-3417},
DOI = {10.3390/app15116110}
}

@article{bommasani2021foundation,
  title={On the Opportunities and Risks of Foundation Models},
  author={Bommasani, Rishi and Hudson, Drew A. and Adeli, Ehsan and Altman, Russ and et al.},
  journal={arXiv preprint arXiv:2108.07258},
  year={2021}
}

@techreport{graefe2016guide,
  title={Guide to Automated Journalism},
  author={Graefe, Andreas},
  institution={Tow Center for Digital Journalism, Columbia University},
  year={2016},
  url={https://www.cjr.org/tow_center_reports/guide_to_automated_journalism.php}
}

@book{diakopoulos2019automating,
  title={Automating the News: How Algorithms Are Rewriting the Media},
  author={Diakopoulos, Nicholas},
  publisher={Harvard University Press},
  year={2019}
}

@inproceedings{kwak2010twitter,
  title={What is Twitter, a Social Network or a News Media?},
  author={Kwak, Haewoon and Lee, Changhyun and Park, Hosung and Moon, Sue},
  booktitle={Proceedings of the 19th International Conference on World Wide Web},
  pages={591--600},
  year={2010},
  organization={ACM}
}

@article{cinelli2021echo,
  title={The echo chamber effect on social media},
  author={Cinelli, Matteo and Morales, G. D. F. and Galeazzi, Alessandro and Quattrociocchi, Walter and Starnini, Michele},
  journal={Proceedings of the National Academy of Sciences},
  volume={118},
  number={9},
  year={2021},
  publisher={National Acad Sciences}
}

@article{zubiaga2018discourse,
  title={Discourse-aware rumour stance classification in social media using sequential classifiers},
  author={Zubiaga, Arkaitz and Aker, Ahmet and Bontcheva, Kalina and Liakata, Maria and Procter, Rob},
  journal={Information Processing \& Management},
  volume={54},
  number={2},
  pages={273--290},
  year={2018},
  publisher={Elsevier}
}

@book{pariser2011filter,
  title={The Filter Bubble: What the Internet Is Hiding from You},
  author={Pariser, Eli},
  publisher={Penguin Press},
  year={2011}
}

@article{gentzkow2006media,
  title={Media bias and reputation},
  author={Gentzkow, Matthew and Shapiro, Jesse M.},
  journal={Journal of Political Economy},
  volume={114},
  number={2},
  pages={280--316},
  year={2006},
  publisher={University of Chicago Press}
}

@article{tversky1974judgment,
  title={Judgment under Uncertainty: Heuristics and Biases},
  author={Tversky, Amos and Kahneman, Daniel},
  journal={Science},
  volume={185},
  number={4157},
  pages={1124--1131},
  year={1974},
  publisher={American Association for the Advancement of Science}
}

@inproceedings{bender2021dangers,
  title={On the Dangers of Stochastic Parrots: Can Language Models Be Too Big?},
  author={Bender, Emily M. and Gebru, Timnit and McMillan-Major, Angelina and Shmitchell, Shmargaret},
  booktitle={Proceedings of the 2021 ACM Conference on Fairness, Accountability, and Transparency (FAccT)},
  pages={610--623},
  year={2021}
}

@inproceedings{sheng2019woman,
  title={The Woman Worked as a Babysitter: On Biases in Language Generation},
  author={Sheng, Emily and Chang, Kai-Wei and Natarajan, Prem and Peng, Nanyun},
  booktitle={Proceedings of the 2019 Conference on Empirical Methods in Natural Language Processing},
  pages={3407--3412},
  year={2019}
}

@inproceedings{goyal2021evaluating,
  title={Evaluating Factuality in Generative Summarization},
  author={Goyal, Tanya and Durrett, Greg},
  booktitle={Proceedings of the 2021 Conference on Empirical Methods in Natural Language Processing},
  pages={178--190},
  year={2021}
}

@article{del2016echo,
  title={Echo Chambers in the Age of Misinformation},
  author={Del Vicario, Michela and Bessi, Alessandro and Zollo, Fabiana and Petroni, Fabio and Scala, Antonio and Caldarelli, Guido and Stanley, H. Eugene and Quattrociocchi, Walter},
  journal={Proceedings of the National Academy of Sciences},
  volume={113},
  number={3},
  pages={554--559},
  year={2016}
}

@book{sunstein2017republic,
  title={\#Republic: Divided Democracy in the Age of Social Media},
  author={Sunstein, Cass R.},
  publisher={Princeton University Press},
  year={2017}
}

@inproceedings{fan2018controllable,
  title={Controllable Abstractive Summarization},
  author={Fan, Angela and Grangier, David and Auli, Michael},
  booktitle={Proceedings of the 2nd Workshop on Neural Machine Translation and Generation},
  pages={45--54},
  year={2018}
}

@inproceedings{spinde2021babe,
  title={Neural Media Bias Detection Using Distant Supervision With BABE},
  author={Spinde, Timo and Westermann, Hendrik and Rehm, Georg},
  booktitle={Proceedings of the 2021 Annual Conference of the North American Chapter of the Association for Computational Linguistics (NAACL)},
  pages={3518--3530},
  year={2021}
}

@inproceedings{pryzant2020automatically,
  title={Automatically Neutralizing Subjective Bias in Text},
  author={Pryzant, Reid and Martinez, Victor and Dass, Nayeon Lee and Jurafsky, Dan and Eisenstein, Jacob},
  booktitle={Proceedings of the 58th Annual Meeting of the Association for Computational Linguistics (ACL)},
  pages={3125--3135},
  year={2020}
}

@article{brown2020language,
  title={Language models are few-shot learners},
  author={Brown, Tom and Mann, Benjamin and Ryder, Nick and Subbiah, Melanie and Kaplan, Jared D and Dhariwal, Prafulla and Neelakantan, Arvind and Shyam, Pranav and Sastry, Girish and Askell, Amanda and others},
  journal={Advances in neural information processing systems},
  volume={33},
  pages={1877--1901},
  year={2020}
}

@article{llm_bias_nature_2024,
  title={Bias of AI-generated content: an examination of news produced by large language models},
  author={Fang, X. and Che, S. and Mao, M. et al.},
  journal={Scientific Reports},
  publisher={Nature},
  year={2024},
  url={https://www.nature.com/articles/s41598-024-55686-2}
}

@article{bias_fairness_survey_2024,
  title={Bias and Fairness in Large Language Models: A Survey},
  author = {Gallegos, Isabel O.  and
      Rossi, Ryan A.  and
      Barrow, Joe  and
      Tanjim, Md Mehrab  and
      Kim, Sungchul  and
      Dernoncourt, Franck  and
      Yu, Tong  and
      Zhang, Ruiyi  and
      Ahmed, Nesreen K.},
  journal={Computational Linguistics},
  publisher={MIT Press},
  year={2024},
  volume={50},
  number={3},
  pages={1097},
  url={https://direct.mit.edu/coli/article/50/3/1097/121961/Bias-and-Fairness-in-Large-Language-Models-A}
}

@misc{reuters_lynx_insight_2018,
  title={Reuters' new automation tool wants to help reporters spot the hidden stories in their data},
  author={Cassidy, Padraic},
  journal={Nieman Journalism Lab},
  year={2018},
  url={https://www.niemanlab.org/2018/03/reuters-new-automation-tool-wants-to-help-reporters-spot-the-hidden-stories-in-their-data-but-wont-take-their-jobs/}
}

@misc{lynx_insight_platform,
  title={Reuters Lynx Insight: Automated Journalism Platform},
author={Meta Guide},
  journal={Meta-Guide.com},
  url={https://meta-guide.com/news/journalism/reuters-lynx-insight}
}

@misc{wapo_heliograf_2016,
  title={The Washington Post experiments with automated storytelling to help power 2016 Rio Olympics coverage},
author={Washington Post},
  journal={The Washington Post},
  year={2016},
  url={https://www.washingtonpost.com/pr/wp/2016/08/05/the-washington-post-experiments-with-automated-storytelling-to-help-power-2016-rio-olympics-coverage/}
}

@misc{heliograf_case_study_2025,
  title={AI Case Study: The Washington Post's Use of AI for Automated Content Creation},
author={Frederik Felipson},
  journal={Redress Compliance},
  year={2025},
  url={https://redresscompliance.com/ai-case-study-the-washington-posts-use-of-ai-for-automated-content-creation/}
}

@misc{fabbri2021,
      title={Multi-Perspective Abstractive Answer Summarization}, 
      author={Alexander R. Fabbri and Xiaojian Wu and Srini Iyer and Mona Diab},
      year={2021},
      eprint={2104.08536},
      archivePrefix={arXiv},
      primaryClass={cs.CL},
      url={https://arxiv.org/abs/2104.08536}, 
}

@misc{yadav2022,
      title={Automatic Text Summarization Methods: A Comprehensive Review}, 
      author={Divakar Yadav and Jalpa Desai and Arun Kumar Yadav},
      year={2022},
      eprint={2204.01849},
      archivePrefix={arXiv},
      primaryClass={cs.CL},
      url={https://arxiv.org/abs/2204.01849}, 
}

@misc{luo2024,
      title={PerSphere: A Comprehensive Framework for Multi-Faceted Perspective Retrieval and Summarization}, 
      author={Yun Luo and Yingjie Li and Xiangkun Hu and Qinglin Qi and Fang Guo and Qipeng Guo and Zheng Zhang and Yue Zhang},
      year={2024},
      eprint={2412.12588},
      archivePrefix={arXiv},
      primaryClass={cs.CL},
      url={https://arxiv.org/abs/2412.12588}, 
}

@misc{zhang2025,
      title={A Comprehensive Survey on Process-Oriented Automatic Text Summarization with Exploration of LLM-Based Methods}, 
      author={Yang Zhang and Hanlei Jin and Dan Meng and Jun Wang and Jinghua Tan},
      year={2025},
      eprint={2403.02901},
      archivePrefix={arXiv},
      primaryClass={cs.AI},
      url={https://arxiv.org/abs/2403.02901}, 
}

@misc{aly2025,
      title={An Evaluation of Large Language Models on Text Summarization Tasks Using Prompt Engineering Techniques}, 
      author={Walid Mohamed Aly and Taysir Hassan A. Soliman and Amr Mohamed AbdelAziz},
      year={2025},
      eprint={2507.05123},
      archivePrefix={arXiv},
      primaryClass={cs.CL},
      url={https://arxiv.org/abs/2507.05123}, 
}

@misc{hayati2024,
      title={How Far Can We Extract Diverse Perspectives from Large Language Models?}, 
      author={Shirley Anugrah Hayati and Minhwa Lee and Dheeraj Rajagopal and Dongyeop Kang},
      year={2024},
      eprint={2311.09799},
      archivePrefix={arXiv},
      primaryClass={cs.CL},
      url={https://arxiv.org/abs/2311.09799}, 
}

@misc{ki2025,
      title={Multiple LLM Agents Debate for Equitable Cultural Alignment}, 
      author={Dayeon Ki and Rachel Rudinger and Tianyi Zhou and Marine Carpuat},
      year={2025},
      eprint={2505.24671},
      archivePrefix={arXiv},
      primaryClass={cs.CL},
      url={https://arxiv.org/abs/2505.24671}, 
}

@misc{jiang2023,
      title={Large-Scale and Multi-Perspective Opinion Summarization with Diverse Review Subsets}, 
      author={Han Jiang and Rui Wang and Zhihua Wei and Yu Li and Xinpeng Wang},
      year={2023},
      eprint={2310.13340},
      archivePrefix={arXiv},
      primaryClass={cs.CL},
      url={https://arxiv.org/abs/2310.13340}, 
}

@misc{powers2025,
      title={The GUS Framework: Benchmarking Social Bias Classification with Discriminative (Encoder-Only) and Generative (Decoder-Only) Language Models}, 
      author={Maximus Powers and Shaina Raza and Alex Chang and Umang Mavani and Harshitha Reddy Jonala and Ansh Tiwari and Hua Wei},
      year={2025},
      eprint={2410.08388},
      archivePrefix={arXiv},
      primaryClass={cs.CL},
      url={https://arxiv.org/abs/2410.08388}, 
}

@misc{ghosh2025,
      title={To Bias or Not to Bias: Detecting bias in News with bias-detector}, 
      author={Himel Ghosh and Ahmed Mosharafa and Georg Groh},
      year={2025},
      eprint={2505.13010},
      archivePrefix={arXiv},
      primaryClass={cs.CL},
      url={https://arxiv.org/abs/2505.13010}, 
}

@inproceedings{ghosh-werner-2026-llm,
    title = "{LLM} {B}ias{S}cope: A Real-Time Bias Analysis Platform for Comparative {LLM} Evaluation",
    author = "Ghosh, Himel  and
      Werner, Nick Elias",
    editor = "Croce, Danilo  and
      Leidner, Jochen  and
      Moosavi, Nafise Sadat",
    booktitle = "Proceedings of the 19th Conference of the {E}uropean Chapter of the {A}ssociation for {C}omputational {L}inguistics (Volume 3: System Demonstrations)",
    month = mar,
    year = "2026",
    address = "Rabat, Marocco",
    publisher = "Association for Computational Linguistics",
    url = "https://aclanthology.org/2026.eacl-demo.19/",
    doi = "10.18653/v1/2026.eacl-demo.19",
    pages = "261--270",
    ISBN = "979-8-89176-382-1",
    }

@inproceedings{Krieger2022, series={JCDL ’22},
   title={A domain-adaptive pre-training approach for language bias detection in news},
   url={http://dx.doi.org/10.1145/3529372.3530932},
   DOI={10.1145/3529372.3530932},
   booktitle={Proceedings of the 22nd ACM/IEEE Joint Conference on Digital Libraries},
   publisher={ACM},
   author={Krieger, Jan-David and Spinde, Timo and Ruas, Terry and Kulshrestha, Juhi and Gipp, Bela},
   year={2022},
   month=jun, collection={JCDL ’22} }

@misc{pryzant2019,
      title={Automatically Neutralizing Subjective Bias in Text}, 
      author={Reid Pryzant and Richard Diehl Martinez and Nathan Dass and Sadao Kurohashi and Dan Jurafsky and Diyi Yang},
      year={2019},
      eprint={1911.09709},
      archivePrefix={arXiv},
      primaryClass={cs.CL},
      url={https://arxiv.org/abs/1911.09709}, 
}

@inproceedings{Vaswani2017,
 author = {Vaswani, Ashish and Shazeer, Noam and Parmar, Niki and Uszkoreit, Jakob and Jones, Llion and Gomez, Aidan N and Kaiser, \L ukasz and Polosukhin, Illia},
 booktitle = {Advances in Neural Information Processing Systems},
 editor = {I. Guyon and U. Von Luxburg and S. Bengio and H. Wallach and R. Fergus and S. Vishwanathan and R. Garnett},
 pages = {},
 publisher = {Curran Associates, Inc.},
 title = {Attention is All you Need},
 url = {https://proceedings.neurips.cc/paper_files/paper/2017/file/3f5ee243547dee91fbd053c1c4a845aa-Paper.pdf},
 volume = {30},
 year = {2017}
}

@misc{lima2025,
      title={Improving RAG Retrieval via Propositional Content Extraction: a Speech Act Theory Approach}, 
      author={João Alberto de Oliveira Lima},
      year={2025},
      eprint={2503.10654},
      archivePrefix={arXiv},
      primaryClass={cs.CL},
      url={https://arxiv.org/abs/2503.10654}, 
}

@misc{logé2025,
      title={Truth Sleuth and Trend Bender: AI Agents to fact-check YouTube videos and influence opinions}, 
      author={Cécile Logé and Rehan Ghori},
      year={2025},
      eprint={2507.10577},
      archivePrefix={arXiv},
      primaryClass={cs.CL},
      url={https://arxiv.org/abs/2507.10577}, 
}

@article{kouris2021,
    title = "Abstractive Text Summarization: Enhancing Sequence-to-Sequence Models Using Word Sense Disambiguation and Semantic Content Generalization",
    author = "Kouris, Panagiotis  and
      Alexandridis, Georgios  and
      Stafylopatis, Andreas",
    journal = "Computational Linguistics",
    year = "2021",
    url = "https://aclanthology.org/2021.cl-4.27/",
    doi = "10.1162/coli_a_00417",
}

@inproceedings{reif2022,
    title = "A Recipe for Arbitrary Text Style Transfer with Large Language Models",
    author = "Reif, Emily  and
      Ippolito, Daphne  and
      Yuan, Ann  and
      Coenen, Andy  and
      Callison-Burch, Chris  and
      Wei, Jason",
    booktitle = "Proceedings of the 60th Annual Meeting of the Association for Computational Linguistics (Volume 2: Short Papers)",
    year = "2022",
    publisher = "Association for Computational Linguistics",
    url = "https://aclanthology.org/2022.acl-short.94/",
    doi = "10.18653/v1/2022.acl-short.94",
    pages = "837--848",
}

@misc{furniturewala2024,
      title={Thinking Fair and Slow: On the Efficacy of Structured Prompts for Debiasing Language Models}, 
      author={Shaz Furniturewala and Surgan Jandial and Abhinav Java and Pragyan Banerjee and Simra Shahid and Sumit Bhatia and Kokil Jaidka},
      year={2024},
      eprint={2405.10431},
      archivePrefix={arXiv},
      primaryClass={cs.CL},
      url={https://arxiv.org/abs/2405.10431}, 
}

@misc{socialchem,
      title={NLPositionality: Characterizing Design Biases of Datasets and Models}, 
      author={Sebastin Santy and Jenny T. Liang and Ronan Le Bras and Katharina Reinecke and Maarten Sap},
      year={2023},
      eprint={2306.01943},
      archivePrefix={arXiv},
      primaryClass={cs.CL},
      url={https://arxiv.org/abs/2306.01943}, 
}

@InProceedings{cmv,
  author =                   {Khalid Al-Khatib and Michael V{\"o}lske and Shahbaz Syed and Nikolay Kolyada and Benno Stein},
  booktitle =                {58th Annual Meeting of the Association for Computational Linguistics (ACL 2020)},
  ids =                      {stein:2020l},
  month =                    jul,
  pages =                    {7067--7072},
  publisher =                {Association for Computational Linguistics},
  site =                     {Seattle, USA},
  title =                    {{Exploiting Personal Characteristics of Debaters for Predicting Persuasiveness}},
  url =                      {https://aclanthology.org/2020.acl-main.632/},
  year =                     2020
}

@misc{wang2023promptinglargelanguagemodels,
      title={Prompting Large Language Models for Topic Modeling}, 
      author={Han Wang and Nirmalendu Prakash and Nguyen Khoi Hoang and Ming Shan Hee and Usman Naseem and Roy Ka-Wei Lee},
      year={2023},
      eprint={2312.09693},
      archivePrefix={arXiv},
      primaryClass={cs.AI},
      url={https://arxiv.org/abs/2312.09693}, 
}

@misc{mu2024largelanguagemodelsoffer,
      title={Large Language Models Offer an Alternative to the Traditional Approach of Topic Modelling}, 
      author={Yida Mu and Chun Dong and Kalina Bontcheva and Xingyi Song},
      year={2024},
      eprint={2403.16248},
      archivePrefix={arXiv},
      primaryClass={cs.CL},
      url={https://arxiv.org/abs/2403.16248}, 
}

@misc{pham2024topicgptpromptbasedtopicmodeling,
      title={TopicGPT: A Prompt-based Topic Modeling Framework}, 
      author={Chau Minh Pham and Alexander Hoyle and Simeng Sun and Philip Resnik and Mohit Iyyer},
      year={2024},
      eprint={2311.01449},
      archivePrefix={arXiv},
      primaryClass={cs.CL},
      url={https://arxiv.org/abs/2311.01449}, 
}

\section*{Ethics and Broader Impact Statement}

\textsc{AutoJourn} is designed to support responsible automated journalism by making the generation process transparent, interpretable, and sensitive to news bias. Nonetheless, the system operates on large language models and social media data, both of which carry ethical considerations. First, extracted perspectives may reflect harmful or extreme viewpoints present in online discussions; while these are surfaced for analysis, they should not be interpreted as endorsements of these views. Second, automated summarisation and news generation systems risk propagating factual errors or subtle framing biases. To mitigate this, \textsc{AutoJourn} includes explicit bias detection and neutralisation components, and allows users to inspect, revise, or reject generated content.

No personal user data is collected by the system, and the demonstration uses publicly accessible web content. We strongly caution against deploying the system in high-stakes editorial contexts without human oversight. Automated news-writing tools can influence public discourse, and their outputs must be critically reviewed to avoid reinforcing misinformation, polarization, or representational harm. We release this system as a research demonstrator to encourage further work on transparency, viewpoint diversity, and ethical safeguards in AI-assisted journalism.

\appendix
\section*{Appendix A: Prompt Designs}

This appendix contains all prompt templates used in the system, including
perspective extraction, RAG-augmented prompting, stance-aware summarisation, and
bias-neutralisation prompts.
\section{Dynamic Topic Modeling Prompt}
\label{dtmprompt}
\begin{lstlisting}[language=Python]
Analyze the following text and provide at least {self.num_topics} most suitable topics, 
along with their percentages and the relevant 15 keywords.
Text:
{preprocessed_doc}
Format the output as a JSON object:
{
 "topics": [
   {
     "name": "Topic1",
     "percentage": 25.0,
     "keywords": ["keyword1", "keyword2", ...]
   }, 
   ...
 ]} 
\end{lstlisting}

\section{Multi-Perspective Extraction Prompts}
\label{mpePrompt}
\subsection{Free-Form Zero-Shot Prompting (Hayati et al.~\cite{hayati2024})}
\begin{lstlisting}[language=Python]
Statement: {statement}
Tell me opinions about the statement as many as possible 
from different people with "Agree" or "Disagree" and 
explain how they have different opinions.

Output should be in the JSON format:
{
  1: {"Stance": "Agree", "Reason": "Write your reason here"},
  2: {"Stance": "Disagree", "Reason": "Write your reason here"}
}
\end{lstlisting}

\subsection{Criteria-Based Zero-Shot Prompting (Hayati et al.~\cite{hayati2024})}
\begin{lstlisting}[language=Python]
Statement: {statement}
Tell me opinions about the statement as many as possible 
from different people with "Agree" or "Disagree", 
one-word or one-phrase criteria that is important 
for their opinions, and explain why.

Output:
{
  1: {"Stance": "Agree", "Criteria": ["example1","example2"],
      "Reason": "Write your reason here"},
  2: {"Stance": "Disagree", "Criteria": ["example1","example2"],
      "Reason": "Write your reason here"}
}
\end{lstlisting}

\subsection{Final Unified Perspective Extraction Prompt (Our Version)}
\begin{lstlisting}[language=Python]
Given the statement: "{statement}", generate at least 6 
opinions with "Agree" or "Disagree" stances, including:
- Criteria (1-2 words) important for the stance
- A single-sentence reason

Output format:
{
  1: {"Stance": "Agree",
      "Criteria": ["c1","c2"],
      "Reason": "Some explanation"},
  2: {"Stance": "Disagree",
      "Criteria": ["c1","c2"],
      "Reason": "Some explanation"}
}
\end{lstlisting}

\section{RAG-Augmented Prompts}
\label{RAGPrompt}
\subsection{Retrieval-Augmented Perspective Generation Prompt}
\begin{lstlisting}[language=Python]
Given the statement: "{statement}",
and the following supporting documents:

### Supporting Documents:
{docs}
### End of Documents.

Generate at least 6 opinions with "Agree" or "Disagree",
including criteria and one-sentence reasons.

Output:
{
  1: {"Stance": "Agree",
      "Criteria": ["example1", "example2"],
      "Reason": "explanation"},
  2: {"Stance": "Disagree",
      "Criteria": ["example1", "example2"],
      "Reason": "explanation"}
}
\end{lstlisting}

\section{Summarisation Prompts}
\label{mpsPrompt}
\subsection{PerSphere Summarisation Prompt (Luo et al.~\cite{luo2024})}
\begin{lstlisting}[language=Python]
Given the query and documents, summarize the perspectives.

Requirements:
1. Include both positive and negative claims.
2. Perspectives must not overlap.
3. Summary must be closely tied to the query.
4. References may exceed one.
Output format is XML.

Query: {query}
Documents: {doc}
\end{lstlisting}

\subsection{Our Prompt for Dual-Perspective Summarisation}
\begin{lstlisting}[language=Python]
Given a statement and a set of perspectives, generate:

1. A summary of agreeing perspectives
2. A summary of disagreeing perspectives

Requirements:
- No overlap between summaries
- Content must remain tied to the original statement

Output (JSON):
{
 "statement": "{statement}",
 "summaries": {
    "agree": "...",
    "disagree": "..."
 }
}
\end{lstlisting}

\subsection{PerSphere Summary Merging Prompt}
\begin{lstlisting}[language=Python]
Given the summarizations, merge them into one summary.
Do not use outside knowledge.

Requirements:
1. Include both positive and negative claims.
2. Merge similar perspectives.
3. Use an XML output format.

Summary: {summaries}
\end{lstlisting}

\subsection{Our Final Summary Merging Prompt}
\begin{lstlisting}[language=Python]
Given the statement and summaries, merge them into one summary.

Requirements:
1. Identify common themes.
2. Ensure both perspectives are equally represented.
3. Use neutral, objective language.
4. Keep the merged summary concise.
5. Include evidence from both summaries.

Output (JSON):
{
 "statement": "{statement}",
 "merged_summary": "..."
}
\end{lstlisting}
\section{News Generation Prompt}
\label{NGPrompt}
\begin{lstlisting}[language=Python]
You are a professional journalist writing for a major news outlet. 
Your goal is to craft a compelling and detailed news article.

Topic: {topic_name}
Summary: {summary}
Key Points and Keywords: {keywords}
Style: {tone_instructions}

Requirements:
Write a long, engaging news article.
Include an attention-grabbing headline at the beginning.
Expand upon the provided summary using the listed keywords. 
Use them naturally throughout the article.
Include historical context, background, or analysis where relevant.
Use the specified style and tone throughout the article.

Instructions:
Start your response with the headline, followed by the full article.
\end{lstlisting}

\section{GPT-4o Evaluation Prompt}
\label{GPTEvalPrompt}
\begin{lstlisting}[language=Python]
You are an impartial judge evaluating multi-perspective summaries.

Criteria:
1. Faithfulness
2. Coverage
3. Distinctiveness
4. Ignore formatting differences

Output:
Rating: [[X]]

Original conversation:
{conversation}

AI Response:
{response}
\end{lstlisting}

\section{Bias Neutralisation Prompts}
\label{BNPrompt}
\subsection{Encoder-Decoder Neutralisation Prompt (Training)}
Used implicitly through fine-tuning on WNC~\cite{pryzant2019}. No
explicit inference prompt required.

\subsection{LLM Safety-Fallback Neutralisation Prompt}
\begin{lstlisting}
The following sentence contains biased language:

{sentence}

Rewrite it in a neutral and objective tone.
\end{lstlisting}

\section*{Appendix B: Evaluation Results}
\label{evalres}
% \begin{table}[H]
% \centering
% \small
% \setlength{\tabcolsep}{8pt}
% \renewcommand{\arraystretch}{1.15}
% \label{tab:semdiv_criteria_combined}
% \begin{tabular}{l
%                 S[table-format=1.4]
%                 S[table-format=1.4]}
% \toprule
% \multirow{2}{*}{Model} & \multicolumn{2}{c}{\textbf{Criteria ($\uparrow$)}}\\
% \cmidrule(lr){2-3}
%  & {\textsc{Social-Chem-101}} & {\textsc{CMV}} \\
% \midrule
% GPT-4             &  0.3919 &  0.3776 \\
% GPT-4o            &  0.3545 &  0.3759 \\
% GPT-3.5           &  0.3100 &  0.2829 \\
% GPT-3 (175B)      &  0.2673 &  0.2046 \\
% Llama3-chat (70B) &  0.3196 &  0.3115 \\
% Mixtral (46.7B)   &  0.3186 &  0.1908 \\
% Zero-shot GPT-4   &  0.2885 &  0.2410 \\
% \midrule
% \rowcolor{gray!10}
% \textbf{ZeroShot: GPT-4o, RAG}   &  \bfseries 0.4410 & \bfseries 0.4484 \\
% \rowcolor{gray!10}
% \textbf{ZeroShot: GPT-4o(No RAG)}&  \bfseries 0.4655 & \bfseries 0.4775 \\
% \bottomrule

% \end{tabular}
% \caption{Semantic diversity (mean cosine distance; higher is better) for \textbf{criteria-based} prompting on \textsc{Social-Chem-101} and \textsc{CMV}. Upper block reproduces values from prior work; lower block (shaded) shows our GPT-4o variants.}
% \end{table}

\begin{table}[H]
\centering
\small
\setlength{\tabcolsep}{6pt}
\renewcommand{\arraystretch}{1.1}

\begin{tabular}{p{2.4cm} 
                S[table-format=1.4]
                S[table-format=1.4]}
\toprule
\multirow{2}{*}{Model} & \multicolumn{2}{c}{\textbf{Criteria ($\uparrow$)}}\\
\cmidrule(lr){2-3}
 & {\textsc{Social-Chem-101}} & {\textsc{CMV}} \\
\midrule
GPT-4                 & 0.3919 & 0.3776 \\
GPT-4o                & 0.3545 & 0.3759 \\
GPT-3.5               & 0.3100 & 0.2829 \\
GPT-3 (175B)          & 0.2673 & 0.2046 \\
Llama3-chat (70B)     & 0.3196 & 0.3115 \\
Mixtral (46.7B)       & 0.3186 & 0.1908 \\
Zero-shot GPT-4       & 0.2885 & 0.2410 \\
\midrule
\rowcolor{gray!10}
\textbf{ZeroShot: GPT-4o \& RAG}     & \bfseries 0.4410 & \bfseries 0.4484 \\
\rowcolor{gray!10}
\textbf{ZeroShot: GPT-4o (No RAG)} & \bfseries 0.4655 & \bfseries 0.4775 \\
\bottomrule
\end{tabular}

\caption{Semantic diversity (mean cosine distance; higher is better) for criteria-based prompting on \textsc{Social-Chem-101} and \textsc{CMV}.}
\label{tab:semdiv_criteria_combined}
\end{table}

% \begin{table}[H]
% \centering
% \begin{tabular}{lccc}
% \hline
% \textbf{Model} & \textbf{BLEU} & \textbf{ROUGE-L} & \textbf{Bias Reduction (\%)} \\
% \hline
% BART & 0.9301 & 0.9653 & 84.7 \\
% T5-Small & 0.9184 & 0.9576 & 82.1 \\
% BART + LLM & 0.9278 & 0.9632 & \textbf{91.5} \\
% \hline
% \end{tabular}
% \caption{Bias neutralization performance on combined WNC + custom test set.}
% \label{tab:neutralization-results}
% \end{table}

\begin{table}[H]
\centering
\small
\setlength{\tabcolsep}{6pt}
\renewcommand{\arraystretch}{1.1}

\begin{tabular}{p{2.6cm}ccc}
\hline
\textbf{Model} & \textbf{BLEU} & \textbf{ROUGE-L} & \textbf{Bias Reduction (\%)} \\
\hline
BART             & 0.9301 & 0.9653 & 84.7 \\
T5-Small         & 0.9184 & 0.9576 & 82.1 \\
BART + LLM       & 0.9278 & 0.9632 & \textbf{91.5} \\
\hline
\end{tabular}

\caption{Bias neutralization performance on combined WNC + custom test set.}
\label{tab:neutralization-results}
\end{table}

\section*{Appendix C: Application Screenshots}
\label{screens}
\begin{figure}[H]
    \centering
    \includegraphics[width=1\linewidth]{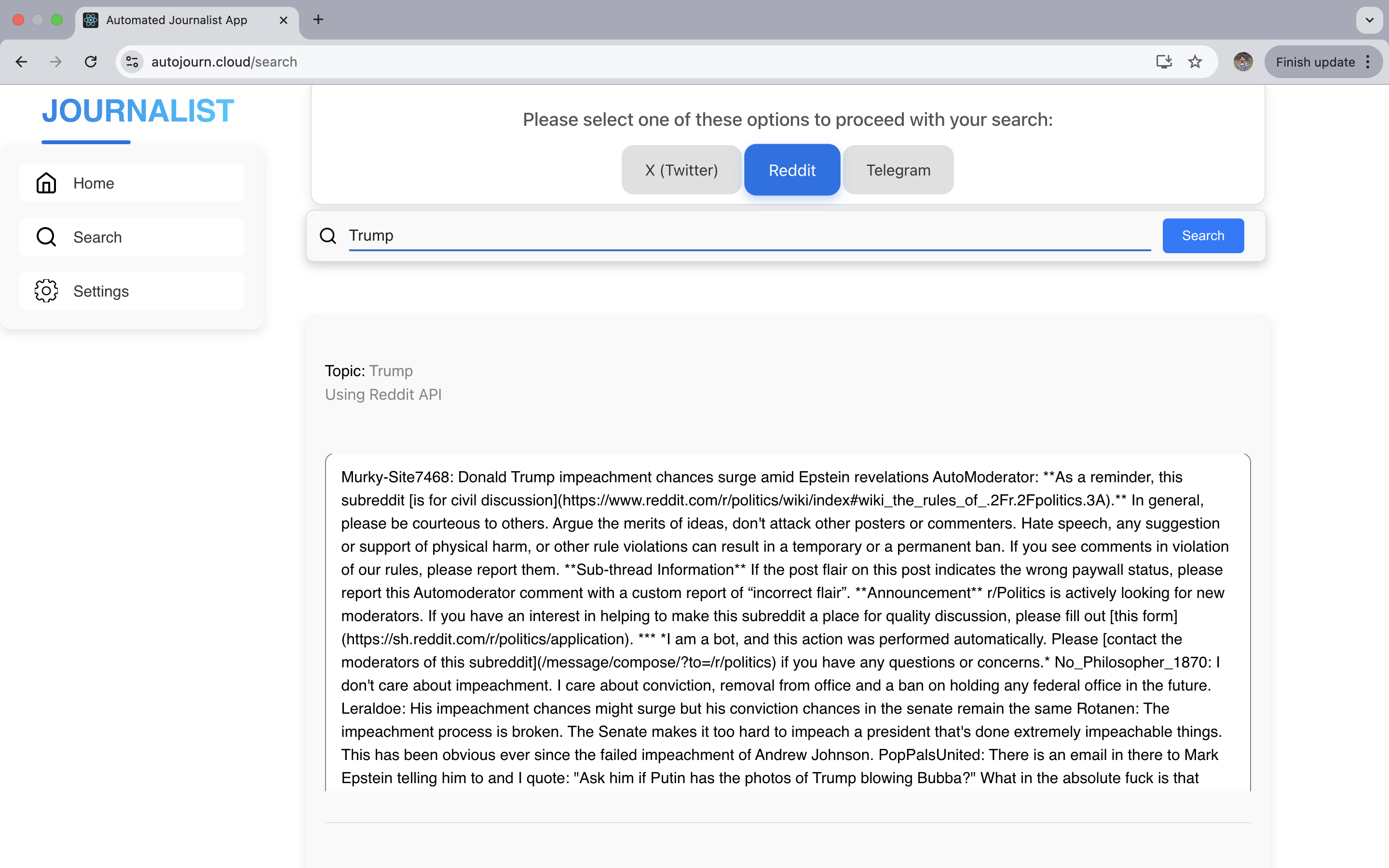}
    \caption{Screenshot of the AutoJourn App from the website. It shows the search results after a query search.}
    \label{fig:autojournss1}
\end{figure}

\begin{figure}[H]
    \centering
    \includegraphics[width=0.75\linewidth]{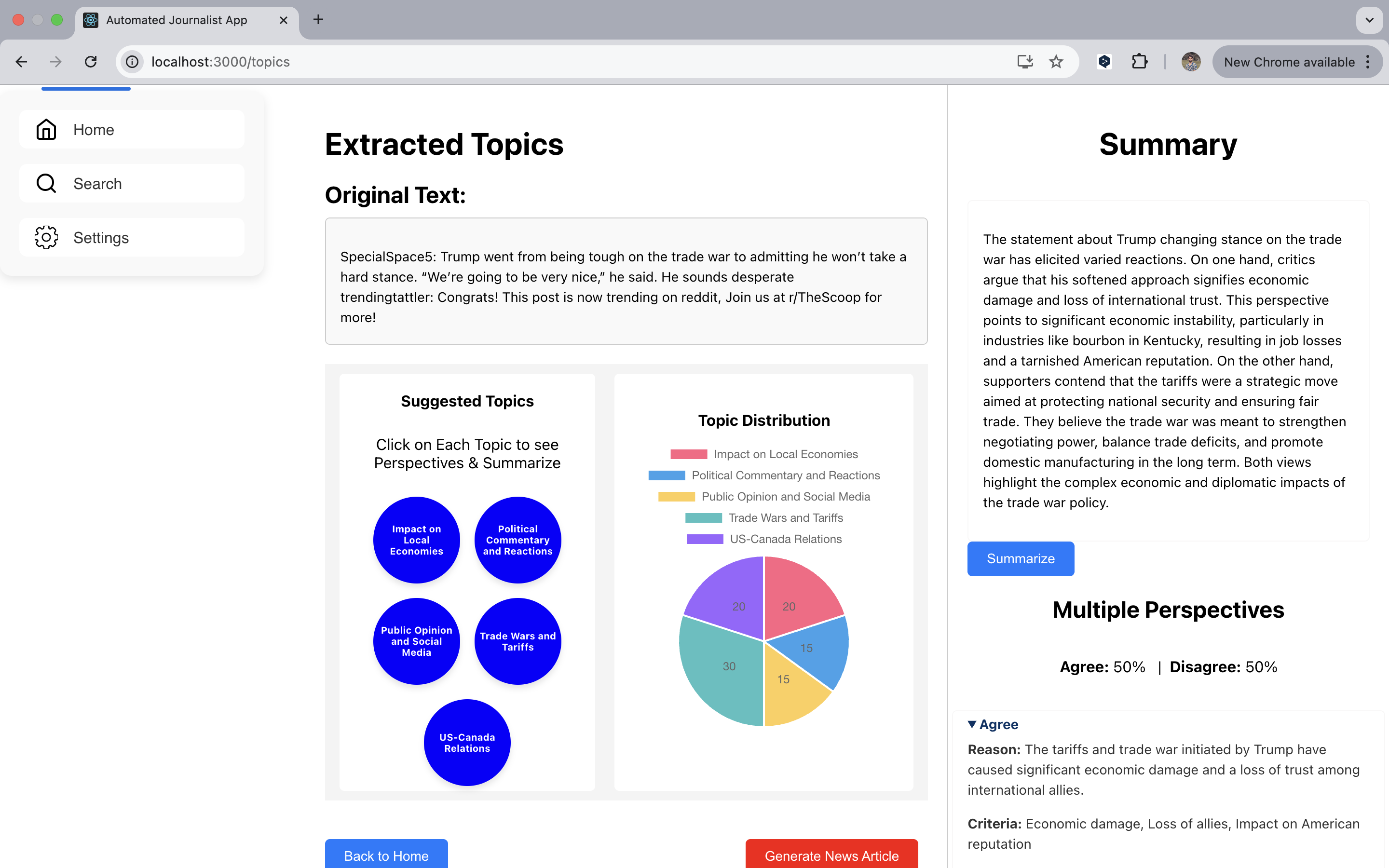}
    \caption{Demonstration of the Multiple Perspectives being displayed on the right-side panel of the AutoJourn Application and the Summarization.}
    \label{fig:multipersp}
\end{figure}

\begin{figure}[H]
    \centering
    \includegraphics[width=0.85\linewidth]{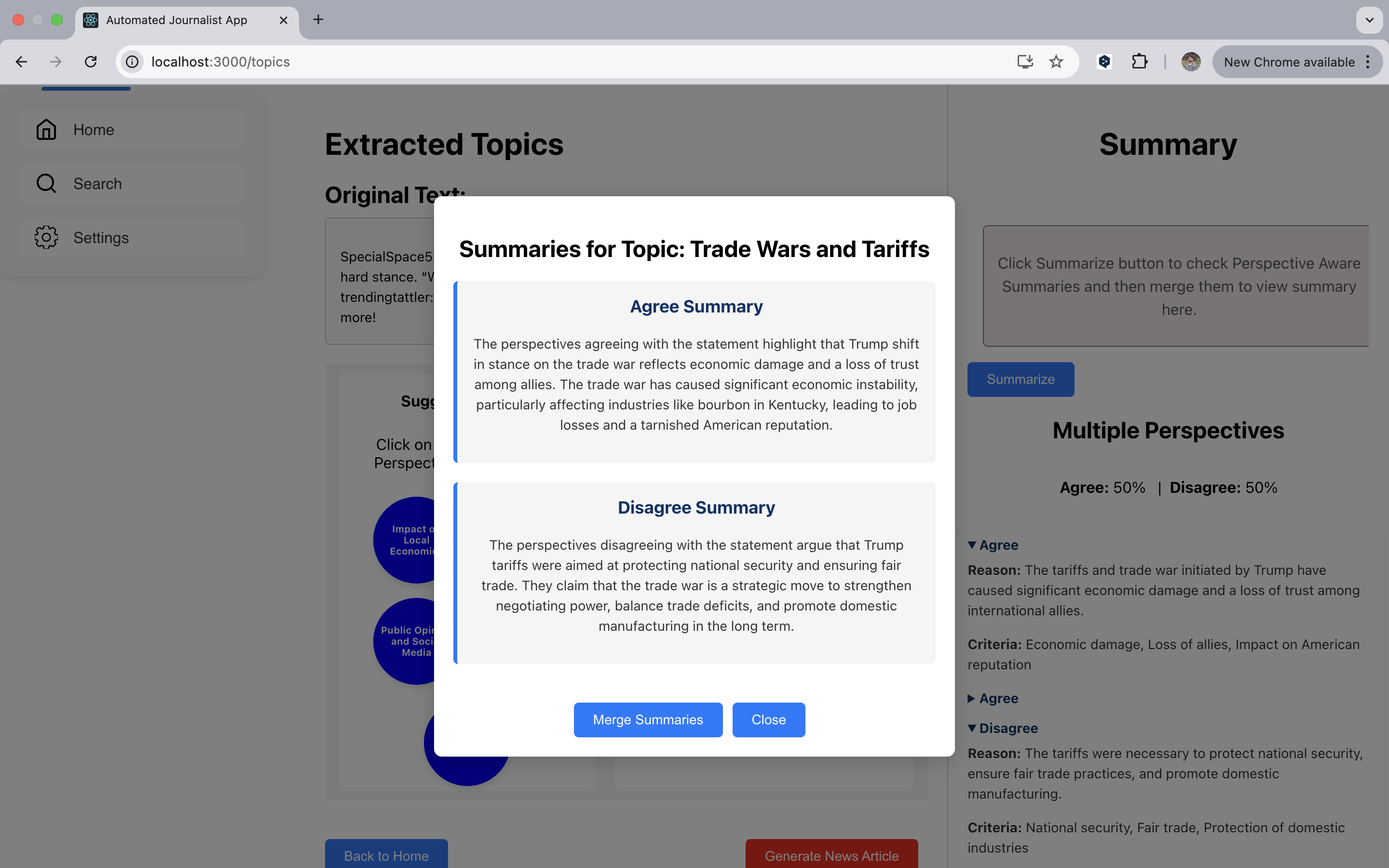}
    \caption{Summarisation module demonstration: shows the agree-disagree stance specific summaries based on a particular topic.}
    \label{fig:summarisation-example}
\end{figure}

\begin{figure}
    \centering
    \includegraphics[width=0.85\linewidth]{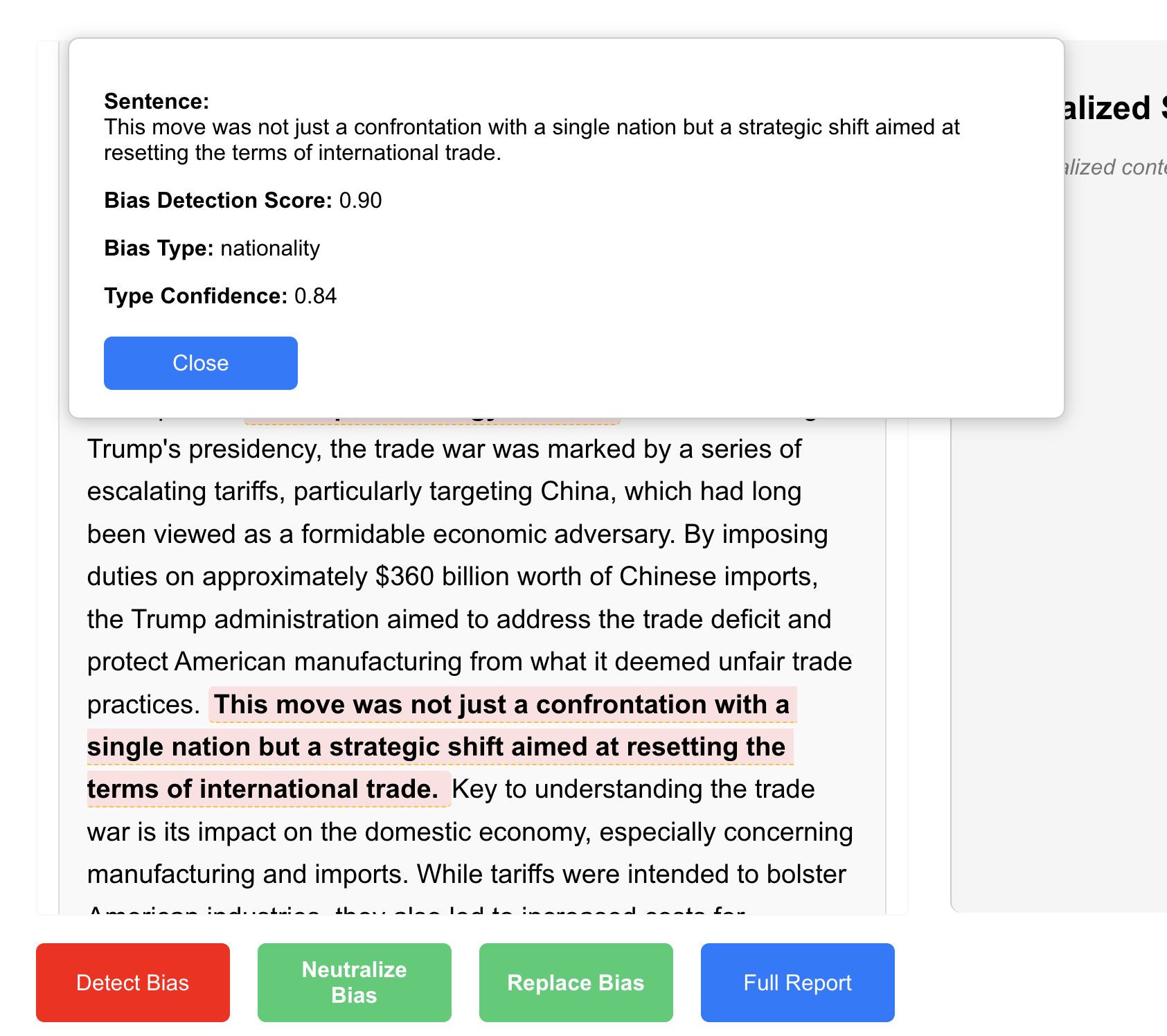}
    \caption{User interface demonstration showing bias-type classification in action within the automated journalist application.}
    \label{fig:ui-bias-type}
\end{figure}

\begin{figure}
    \centering
    \includegraphics[width=0.85\linewidth]{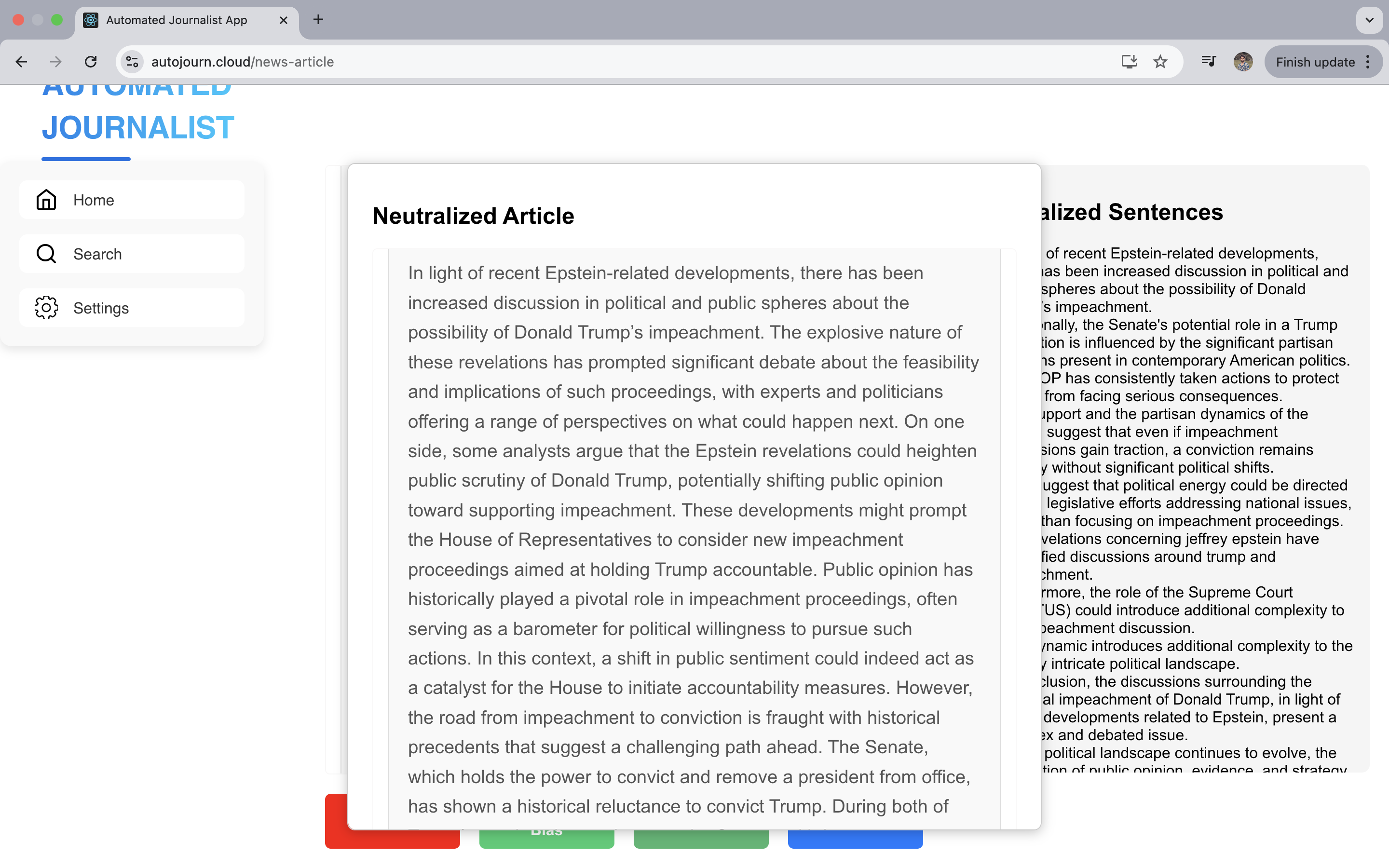}
    \caption{Example of bias neutralization in our system: the biased sentence is rewritten into a neutral form while preserving factual accuracy.}
    \label{fig:neutralization-example}
\end{figure}

\end{document}